\documentclass[conference]{IEEEtran}
\IEEEoverridecommandlockouts
\usepackage{cite}
\usepackage{amsmath,amssymb,amsfonts}
\usepackage{graphicx}
\usepackage{textcomp}
\usepackage{xcolor}
\usepackage{url}
\usepackage{multirow}
\usepackage{bbm}
\usepackage{booktabs}
\usepackage[linesnumbered,ruled,vlined]{algorithm2e}
\SetKwInput{KwInput}{Input}
\SetKwInput{KwOutput}{Output}
\SetKwInput{KwFunction}{Function}
\makeatletter
\newcommand*{\rom}[1]{\expandafter\@slowromancap\romannumeral #1@}
\usepackage[linesnumbered,ruled,vlined]{algorithm2e}
\makeatother
\renewcommand{\baselinestretch}{1}

\begin{document}

\title{Evaluating Distribution System Reliability with Hyperstructures Graph Convolutional Nets 
}

\def\plainauthor{\IEEEauthorblockN{Yuzhou Chen\textsuperscript{\rm 1,2},
Tian Jiang\textsuperscript{\rm 3},
        Miguel Heleno\textsuperscript{\rm 2},
        Alexandre Moreira\textsuperscript{\rm 2},
        Yulia R. Gel\textsuperscript{\rm 3,2}}}
\author{\plainauthor\\ \IEEEauthorblockA{
{$^{1}$Department of Computer and Information Sciences, Temple University, Philadelphia, USA} \\ 
{$^{2}$Energy Storage and Distributed Resources Division, Lawrence Berkeley National Laboratory, Berkeley, USA} \\
{$^{3}$Department of Mathematical Sciences, The University of Texas at Dallas, Dallas, USA}\\
}}

\maketitle
\begin{abstract}
Nowadays, it is broadly recognized in the power system community that to meet the ever expanding energy sector’s needs, it is no longer possible to rely solely on physics-based models and that reliable, timely and sustainable operation of energy systems is impossible without systematic integration of artificial intelligence (AI) tools. Nevertheless, the adoption of AI in power systems is still limited, while integration of AI particularly into distribution grid investment planning is still an uncharted territory. We make the first step forward to bridge this gap by showing how graph convolutional networks coupled with the hyperstructures representation learning framework can be employed for accurate, reliable, and computationally efficient distribution grid planning with resilience objectives. We further propose a Hyperstructures Graph Convolutional Neural Networks (Hyper-GCNNs) to capture hidden higher order representations of distribution networks with attention mechanism. Our numerical experiments show that the proposed Hyper-GCNNs approach yields substantial gains in computational efficiency compared to the prevailing methodology in distribution grid planning and also noticeably outperforms seven state-of-the-art models from deep learning (DL) community.
\end{abstract}


\section{Introduction}
The most recent cases of cyber-physical failures and natural disasters, such as 2021 Texas power crisis, have caused billions of dollars of irreparable damage in grid assets and long-term interruption of service. As a result, there is raising governmental awareness regarding resilience of energy infrastructures~\cite{us_pres} to face such High Impact Low Probability (HILP) events. From a power grid perspective, these events are different from normal routine failures and are often focused on the tails of outage distributions, describing resilience as a risk metric, such as Value at Risk (VaR) or Conditional Value at Risk (CVaR), associated with the loss of load~\cite{Poudel2019}. 

Stochastic optimization is arguably the most popular approach to solve power distribution grid expansion and planning problems that explicitly models extreme events~\cite{Nazemi2020,  Lagos2020}. Although these stochastic optimization approaches allow for assessing resilience while determining investment in expansion plans, they entail a significant computational burden. Particularly in the case of {\it distribution networks}, composed by a large number of nodes and multiple small candidate assets, the number of variables and constraints of stochastic programs dramatically increase, creating a barrier for the application of accurate resilience planning to real systems. As a result, it is of critical importance to develop
new techniques enabling us to produce accurate and feasible expansion plans in the context of real-scale utility networks. This opens a new opportunity for AI and, especially, automatic classification algorithms. Indeed, such AI approaches might be able to significantly improve the computational time of the distribution grid planning process, just by deriving relevant topological information from pre-computed plans, i.e., without requiring a detailed modeling of the intra-hour stochastic events. These classification methods, particularly, based on the graph convolutional network (GCN) architectures,
already proved to be efficient in bioinformatics, social sciences, and transportation systems~\cite{wu2020comprehensive,liu2020towards,zhang2020deep}.
Furthermore, most recently GCNs have emerged into the analysis of power transmission systems~\cite{lonapalawong2022interpreting, mansourlakouraj2022multi, nauck2022predicting}. However, the capability of these classification approaches to reproduce {\it distribution grid resilience} metrics (such as the CVaR of the loss of load) is still an open question and to the best of our knowledge, GCNs have never been used for assessment of distribution grids and the associated distribution expansion plans.

We propose a novel resilience classification approach for accurate, reliable, and computationally efficient distribution grid planning called Hyperstructures Graph Convolutional Neural Networks (Hyper-GCNNs). Hyper-GCNNs is based on the convolutional architecture with a hyperstructure learning mechanism and innovative graph-theoretic summaries of distribution grid vulnerability. We then evaluate how well Hyper-GCNNs captures the overall resilience of the expansion plans and mirror conclusions of the substantially more expensive CVaR metric. (Our expansion plans here are constructed using stochastic simulation with the optimal distribution network operation as adopted in operational planning.) {Hyper-GCNNs ranks the potential power expansion plans in terms of their resiliency while using only the higher-order network information and bypassing costly optimization. Such classification results can be then employed by utilities and regulators to enhance their understanding how each expansion plan changes the operational risk of the overall system, to pre-select the most promising plan candidates, and more generally, to develop pro-active risk mitigation policies.}

{\bf Beyond Distribution Grids} While here we primarily focus on data mining for distribution grid investment planning, the proposed Hyper-GCNNs methodology is broadly applicable to many other diverse networks which exhibit multimodality of nodes and intrinsic heterogeneity in higher-order interactions, such as crypto-exchanges and individual addresses or transportation hubs, urban and rural stations.

The key novelty of this paper are summarized as follows:
\begin{itemize}
    \item This is the first attempt to introduce the concepts of GDL, hypergraph learning, and data mining to the yet uncharted territory
of {\it distribution grid investment planning}.
\item We introduce a new graph-theoretic summary, {\it uniqueness scores (U-scores)}
which allows for exploiting the efficiency of information flow among different types of nodes such as loads and substations in distribution grids or hubs and rural areas in transportation systems.

\item We develop a new Hyperstructures Graph Convolutional Neural Networks (Hyper-GCNNs) which enable for systematic characterization of the latent higher order network representations with an attention mechanism.
Our experiments on the validation of 
Hyper-GCNNs in conjunction with distribution grid planning indicates that Hyper-GCNNs delivers reliable classification of power expansion plans while exhibiting highly competitive computational efficiency.
    
\end{itemize}

\section{Related Work}
{\bf Artificial Intelligence Tools for Power System Analysis}  AI tools and, particularly, various machine learning (ML) and deep learning (DL) approaches start to propagate into a broad range of power system studies~\cite{zhao2019artificial, jufri2019state, ibrahim2020machine}. Some of the most prominent recent examples are {\it L2RPN: Learning to Run a Power Network in a Sustainable World}~\cite{marot2020l2rpn} under NeurIPS2020 and {\it The Learning to run a power network} challenge under at NeurIPS2018~\cite{Apogee}, both largely focusing on the utility of reinforcement learning for smart-grid operations.
Other recent examples are Bayes networks and meta-action for cascading failure propagation~\cite{pi2018machine, huang2019cascading}, node classification in transmission power grids using graph neural networks (GNNs)~\cite{LFGCN}, semi-supervised learning for load monitoring and scheduling~\cite{gillis2016non},
fault detection using Support Vector Machine (SVM) and other ML models~\cite{eskandarpour2017improving,  jenssen2019intelligent}, as well as topology-based systems as a part of neural network architectures for impact metrics classification and prediction in power systems~\cite{NREL_IAAI21}. However, comparing to other knowledge domains, the AI methodology still remains severely underused in the power system analysis~\cite{warren2019can,ibrahim2020machine, you2020review} and most existing DL approaches primarily focus on power transmission system~\cite{huang2020recurrent,liu2021searching, beyza2022characterising, lonapalawong2022interpreting, mansourlakouraj2022multi, nauck2022predicting}. To the best of our knowledge, AI and DL, in particular, have neither been used before to assess resilience of potential distribution grid investment plans, nor with distribution grid planning.
\begin{figure}[h]
\centering
\includegraphics[width=0.49\textwidth]{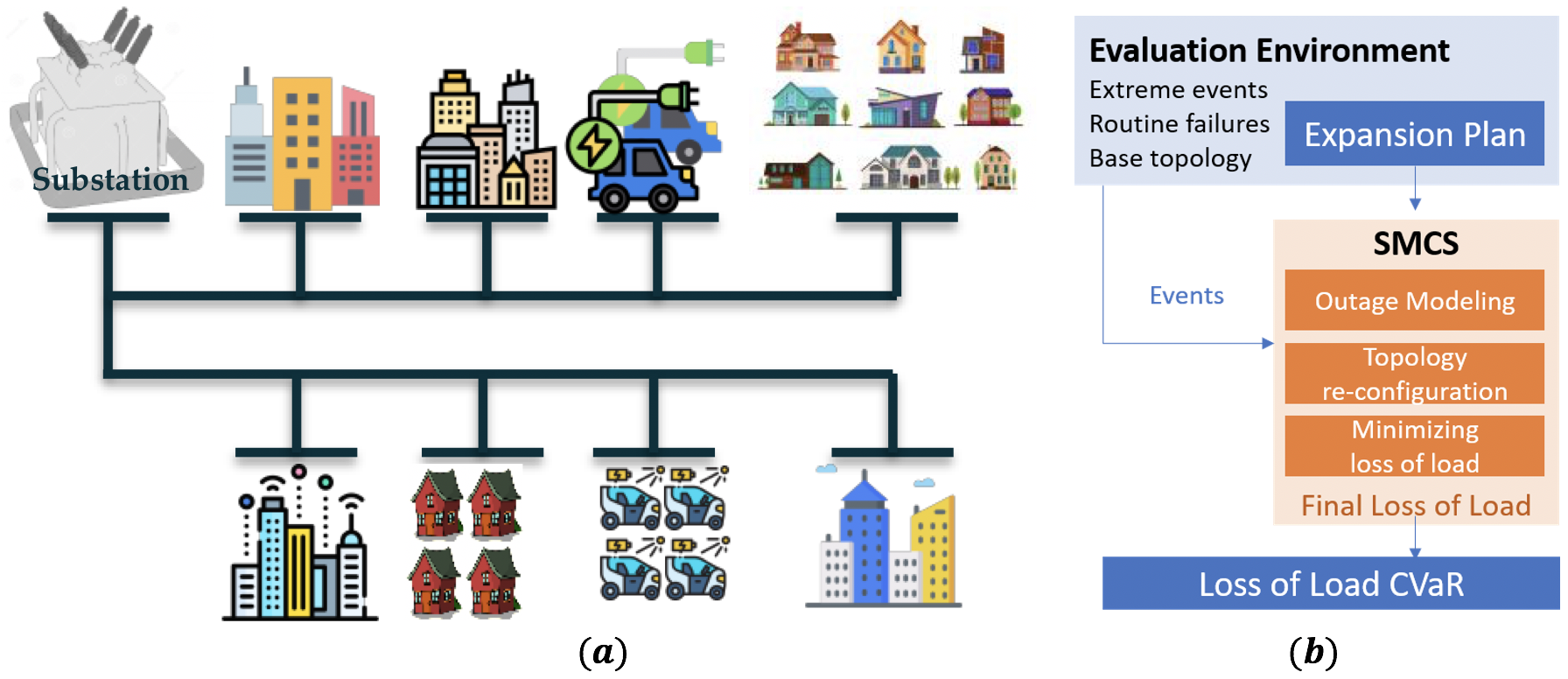}
\caption{{(a) General distribution system, containing the substation (main power supply), the network lines and the consumption nodes, generally represented by multiple buildings and (b) base resilience evaluation methodology, in which the expansion and planning solution is tested against multiple scenarios of network outages caused by either routine failures or by extreme weather events. Each outage scenario leads to a loss of load which entails significant costs to the system. The conditional value at risk (CVaR) of this cost is a widely-used  resilience metric in the planning-stage.}
\label{fig:generalDistributionSystem}}
\end{figure} 

{\bf Methods for Distribution Grid Planning} Current approaches in distribution grid planning heavily rely on mathematical programming frameworks. For instance, \cite{Munoz2018} formulates a distribution grid planning as a mixed-integer optimization problem that seeks to identify the optimal portfolio of investments in lines, transformers and substations in order to minimize costs associated with expected energy not supplied. The approach developed by~\cite{Munoz2016} presents an optimization model to determine generation and network expansion across a given period of time while considering reliability. In turn, \cite{Nazemi2020} proposes a linear programming based approach for determining investments in storage devices in order to improve the resilience of the distribution system against earthquakes. Recently, \cite{Ma2019} develops a mixed-integer stochastic model to optimize investments in line hardening, distributed generation and line switches so as to decrease costs related to expected loss of load and operation of distribution grids. The majority of optimization problems associated with distribution grid planning belong to the {mixed-integer linear  programming} (MILP) class, which is well-known as an NP-Hard class. There are many ways to facilitate (e.g., decomposition techniques) the solution of MILP problems, especially when there are too many variables and constraints involved as in the case of planning problems. However, all of these ways (in classic optimization) involve the utilization of the Branch and Bound algorithm, whose computational complexity is exponential. Hence, depending on the size of the distribution system and the number of candidate assets, the expansion problem may take a prohibitive time to be solved. Adequate formulation of the distribution planning problem depends on the definition of a metric to be improved. Traditional distribution grid planning focuses on reliability indices, which are created to define routine failures of the power grid~\cite{Billinton1989}, and later included into existing optimization expansion and planning methods for the distribution network~\cite{Munoz2016}. Here, instead of reliability, we are interested in assessing the resilience of the system by estimating the CVaR of the loss of load given a defined expansion plan. To do so, unlike the conventional manner, we propose to replace conventional optimization by a DL-based method so as to compute our metric in a computational efficient way.  

\textbf{Hypergraph Neural Networks}
Hypergraph deep learning is a newly emerging direction in GNNs which allows us to address various complex higher order interdependencies (i.e., beyond node pairwise interactions). For instance, HGNN~\cite{Feng2019aaai} is arguably the first attempt to generalize GNNs to hypergraph learning. In turn,~\cite{yadati2019hypergcn} proposes an alternative weighting mechanism to derive hypergraph Laplacian via mediators. In spirit of GraphSAGE, HyperSAGE of~\cite{arya2020hypersage} uses a spatial approach to study message propagation in and between hyperedges.
Most recently, \cite{Huang2021ijcai} compares existing graph and hypergraph neural networks, and formulates a unified framework for hypergraph learning. Finally, \cite{chien2022iclr} investigates tensor-based propagation on $d$-uniform hypergraphs.
To the best of our knowledge, hypergraph learning has never been applied to power systems and, distribution grids, in particular.

\section{Methodology}

The broader objective of this paper is to investigate utility of DL-based classification methods to evaluate power distribution grid expansion plans in terms of their resilience performance. In particular, our analysis aims to assess the ability to approximate explicit but computationally expensive risk-based resilience metrics, such as the CVaR of loss of load, through the topological descriptors of the distribution networks. 
That is, our ultimate idea is to gather different distribution expansion plans and to compare their resilience evaluation, based on the prevailing CVaR calculations, with the new topology-based metrics. 
In this section, we present (i) an overview of the traditional risk-based resilience evaluation methods used for comparison in our analysis, (ii) the algorithm to obtain the new topological signature of distribution grid networks, i.e., {\it uniqueness scores}, and (iii) our proposed Hyper-GCNNs designed for graph learning.

For the purpose of modeling distribution grids (as in the stylized representation of Fig.~\ref{fig:generalDistributionSystem} (a)), we consider a graph $\mathcal{G}=(\mathcal{V}, \mathcal{E}, \omega)$ with a node set $\mathcal{V}$, an edge set $\mathcal{E}$, and (edge)-weighted function $\omega: \mathcal{E} \mapsto \mathbb{R}_{\ge 0}$. Number of nodes $|\mathcal{V}|=N$ and distance among two nodes in $\mathcal{G}$ is denoted by $d_{uv}$, with $d_{uv}\equiv 0$ if there exists no path connecting nodes $u$ and $v$. Let $A \in \mathbb{R}^{N \times N}$ denote a symmetric adjacency matrix with $N$ nodes, and $D \in \mathbb{R}^{N \times N}$ be a diagonal matrix with $D_{uu} = \sum_{v} A_{uv}$. Here, nodes represent buses and substations; edges reflect distribution lines, and the (edge)-weighted function
is used to model power flow information about the system. In contrast to currently existing approaches based on graph-theoretical analysis of power grid networks that do not distinguish among types of nodes, we make a {\it clear} distinction between substations (nodes of the distribution system that serve as a connection to the main transmission grid) and buses (nodes of the distribution system that may have a power load). 

{\bf Mathematical Formulation of the Problem} Our targets are to (i) learn a mapping function $\mathfrak{F}_c: \mathcal{G} \rightarrow \mathcal{Y}_c$ that classify distribution grids into different classes (where $\mathcal{Y}_c$ is the label of distribution network $\mathcal{G}$) and (ii) find a function $\mathcal{F}_r$ to predict CVaR value of distribution grid based on graph structural information and node features, i.e., $\mathfrak{F}_r: \mathcal{G} \rightarrow \mathcal{Y}_r$ (where $\mathcal{Y}_r$ is CVaR value of distribution network $\mathcal{G}$).

\subsection{Base Resilience Evaluation Methodology} \label{sec:BaseResilienceEvaluationMethod}

Reliability and resilience evaluation methods aim at assessing the ability of the power grid to supply the load, considering multiple scenarios of grid outages caused by routine failures of equipment or by High Impact Low Probability (HILP) events, such as storms, earthquakes, wildfires, and cyber-physical attacks. Here, there is a distinction between reliability and resilience metrics: the reliability performance is associated with the routine outages and, therefore, with the expected values of loss of load as well as frequency and duration of the interruptions; in contrast, resilience is more associated with the ability of the grid to withstand and mitigate long-duration outages caused by HILP events, which means that the tail aspects (risk) of the loss of load is the key for the resilience performance. Hence, different power grid expansion plans with specific circuits connections, cables and lines parameters, substation locations, etc. – lead to different risks of loss of load, which means these resilience evaluations are essential to support electric utilities planning decisions. 

Considering an evaluation environment with a pre-defined set of scenarios of routine failures and HILP events, the objective of the resilience evaluation is to quantify the distribution of the loss of load on the different nodes of the grid together with corresponding CVaR of the distribution. This is typically achieved via Sequential Monte Carlo Simulation (SMCS) methods~\cite{heydt2010distribution}, which perform simulations of loss of load over annual time horizons. More specifically, {to assess the annual CVaR of loss of load according to the orange box in Fig.~\ref{fig:generalDistributionSystem} (b), we simulate operation of the distribution system for several scenarios (e.g., 2000 scenarios). Here each scenario corresponds to 365 days (one year) and the system is operated for each hour of each day. Due to its respective failures, each annual scenario is associated with the annual loss of load. Given the amounts of loss of load for all scenarios, we then compute the annual CVaR of loss of load. Such a procedure can take up to 24 hours for a 54-bus system (see the case study in Section~\ref{exp}). Hence, one of the primary motivations of this paper is to replace the costly simulations related to the orange box in Fig.~\ref{fig:generalDistributionSystem} (b) by a computationally efficient ML/DL tool 
that can quickly classify distribution grids according to their corresponding ranges of annual CVaR of loss of load.}

\subsection{New Uniqueness Scores}

Conventional resilience metrics for power grid networks based on graph-theoretic foundations are average path length (APL), average betweenness centrality (Avg. BC), diameter ($\mathfrak{D}$), small-worldness and giant component properties~\cite{
abedi2019review}. That is, network summaries which first, do not differentiate among types of nodes in the power grids (i.e., substations vs. buses) and, second, largely address only global network characteristics and disregard important information on local graph topology. 

Our goal is to define a new node feature for load based on the simple paths between load and substation. Inspired by the notion of graph APL, we propose a new network summary, namely,  {\it uniqueness scores (U-scores)}, in order to exploit the efficiency of information (i.e., flow) transport between different loads and substations. We now summarize our U-scores algorithm in a framework for different potential expansion plans of a distribution system. The U-scores framework has three main components: (i) {\bf paths} extraction, (ii) {\bf scores} calculation, and (iii) {\bf uniqueness scores} finalization. We discuss each component in details as follows.

Given the original graph structure $\mathcal{G}\textsubscript{Bus}$ and the expansion plan $\mathcal{G}\textsubscript{Exp}$, $\mathcal{G}\textsubscript{Comb}$ represents the combination of $\mathcal{G}\textsubscript{Bus}$ and $\mathcal{G}\textsubscript{Exp}$. Let $l_j$ and $s_i$ denote the load node $j$ and substation node $i$ in $\mathcal{G}\textsubscript{Comb}$, respectively. We then define the new U-scores through the steps below: 
\begin{enumerate}
    \item {\bf Paths}. We compute the list of simple weighted paths from $l_j$ and $s_i$, i.e., $\textnormal{simplePaths}_{l_j,s_i}$; then we sort $\textnormal{simplePaths}_{l_j,s_i}$ in increasing order by summarizing the edge weights in path, and the sorted $\textnormal{simplePaths}_{l_j,s_i}$ is defined as $\textnormal{simplePaths}^{\text{sorted}}_{l_j,s_i}$.
    
    \item {\bf Scores}. Suppose there are $\mathcal{K}$ (i.e., $\textnormal{len}(\textnormal{simplePaths}^{\text{sorted}}_{l_j,s_i})$) paths in $\textnormal{simplePaths}^{\text{sorted}}_{l_j,s_i}$, we first assign the uniqueness score of the shortest weighted path $S^{(1)}_{l_j, s_i}$ as 1, use a list $U_{l_j, s_i}$ to store the corresponding simple path by union operation, and calculate its length. For clarity, we use $P_{l_j, s_i}$ to record the total length of simple paths in $\textnormal{simplePaths}^{\text{sorted}}_{l_j,s_i}$. For $k$-th path $\textnormal{simplePaths}^{\text{sorted}}_{l_j,s_i}[k]$ (where $2 \leq k \leq \mathcal{K}$), the calculation of uniqueness score involves two extra sub-steps: (i) generate the intersection $I_{l_j, s_i}$ between $\textnormal{simplePaths}^{\text{sorted}}_{l_j,s_i}[k]$ and stored $U_{l_j, s_i}$, indicating that the repetition rate of $k$-th path; (ii) calculate the {\it normalized} uniqueness score based on the ratio of the length of the intersection to the length of $k$-th path.
    
    \item {\bf Uniqueness score}. We finalize the uniqueness score of load $l_j$ on substation $s_i$ as the ratio between the summation of scores and total length of paths in $\textnormal{simplePaths}^{\text{sorted}}_{l_j,s_i}$. 
\end{enumerate}

In our U-scores algorithm, we first encode path information of the distribution system using scores for the load in the system, since traditional topological metrics like~\cite{chanda2016defining} usually assess robustness of power system network in graph-level. U-scores reduce computational costs of paths extraction and only aggregate information between load and substation. The basic idea is that, graph-level resiliency metrics are actually not useful to capture the relationship between load and substation, since they only provide very global and limited structural information.

{\bf What new do U-scores bring?} In addition to accurately evaluating  robustness of distribution systems, the new U-scores metric sheds a valuable insight into {\it how each load affects the robustness of distribution system complex network}. Such quantitative evaluation yields multi-fold benefits. First, it
allows power system operators to enhance their understanding of the load dynamics and, hence, to develop more efficient and pro-active risk mitigation strategies.
Second, armed with the U-scores metric, electrical engineers can evaluate 
potential expansion plans of the existing distribution systems in a more accurate and reliable manner via a comparison on U-scores summation $\sum\sum{\boldsymbol{S}}$ (i.e., higher values indicate more resilient distribution system).

Note that the currently existing graph-theoretical summaries 
{\bf do not allow} for such a comparison of expansion plans. For instance,
Table~\ref{metric_comparsion} shows the conventional graph-theoretic summaries of power grid vulnerability (i.e., APL, $\mathfrak{D}$, and Avg. BC), the proposed new U-scores, and CVaR. Note that lower APL, $\mathfrak{D}$, Avg. BC, CVaR~\cite{
abedi2019review} and higher U-scores are preferred and considered as indicating higher system resilience.  Here, APL, $\mathfrak{D}$, Avg. BC and U-scores are based only on the topology of the considered distribution networks, while CVaR is based on a stochastic optimization approach and explicitly accounts for both network topology and power flow information.
As such, CVaR conclusions on resiliency of each plan are viewed by power system analysts as the most reliable results, that is, ``ground truth'' in our context. Remarkably, as Table~\ref{metric_comparsion} suggests conclusions of APL, $\mathfrak{D}$ and Avg. BC are in contradiction with CVaR, while the new U-scores, which uses substantially less information than CVaR, and is noticeably more computationally efficient and has highest correlation with CVaR. More specifically, as shown in Table~\ref{metric_comparsion}, we compute correlation coefficients ($r$) between graph-based metrics and the inverse of CVaR values. We find that (i) Avg. Degree vs. 1/CVaR: $r = 0.869$, 
(ii) Avg. BC vs. 1/CVaR: $r = 0.282$, 
(iii) APL vs. 1/CVaR: $r = -0.133$,
(iv) $\mathfrak{D}$ vs. 1/CVaR: $r = -0.205$, and (v) {\bf U-scores vs. 1/CVaR}: $r = 0.893$. This phenomenon can be explained by the fact that in contrast to conventional APL, $\mathfrak{D}$ and Avg. BC metrics for power system resilience, U-scores explicitly differentiates among substations and load buses and accounts for the local topology of the distribution power grid.

\begin{table}[h!]
\centering
\caption{Reliability metrics for 
systems \# 1 and \# 65.\label{metric_comparsion}}
\begin{tabular}{lcc}
\toprule
\textbf{Metric} & \textbf{Expanded system \# 1} & \textbf{Expanded system \# 65} \\
\hline
APL &5.79&7.00\\
$\mathfrak{D}$ &19.96&24.41\\
Avg. BC &45.37&103.70\\
\midrule
CVaR &28679.76&16736.37\\
U-scores &25.32&43.01\\
\bottomrule
\end{tabular}
\end{table}

\subsection{Hyperstructures}
{A number of recent results indicate that higher order substructures may play important roles in resiliency and functionality of cyber-physical systems and, in particular, power grids~\cite{guedes2016non, dey2019network,skardal2020higher}. This phenomenon can be explained by the intrinsic higher order relations
in power distribution networks, where buses interact with each another, and the power flow through branches may be viewed as a result of such interactions.  
One of the emerging approaches to address such higher order properties is via hypergraph representation learning.   Inspired by these recent results on hypergraph modeling, we propose 
to represent high-order interactions on distribution grids via hyperstructures. That is, we use similarity information about groups of nodes (or edges) instead of only graph connectivity information such as network motifs. We introduce the details of proposed hyperstructures for distribution grid representation learning as follows.} In the graph $\mathcal{G}=(\mathcal{V},\mathcal{E},\omega)$, each edge is incident to two nodes. 
Hypergraph can be viewed as a generalized graph that describes high-order interactions in the data. 
It introduces a structure -- hyperedge, which can contain an arbitrary number of nodes.
Let $\mathcal{H}_{\mathcal{G}}=(\mathcal{V},\mathcal{H}_{\mathcal{E}},\omega_{\mathcal{E}})$ be a hypergraph, where $\omega_{\mathcal{E}}$ represents the hyperedge weight. Then $\mathcal{G}$ is a special case when each hyperedge consists of exactly two nodes, a 2-uniform hypergraph (hyperedge degree $\delta(h_{\mathcal{E}})=|h_{\mathcal{E}}|=2$ for any  $h_{\mathcal{E}} \in \mathcal{H}_{\mathcal{E}}$).
Similarly, we define incidence matrix $Q_{\mathcal{E}}$ for connectivity between node and hyperedge, i.e., $q_{\mathcal{E}}(v,h_{\mathcal{E}}) = 1$ if node $v$ is incident to hyperedge $h_{\mathcal{E}}$ and 0 otherwise.
Then the normalized hypergraph Laplacian is $L_{\mathcal{E}} = I - \frac{1}{2}D^{-1/2}Q_{\mathcal{E}}W_{\mathcal{E}}D_{\mathcal{E}}^{-1}Q_{\mathcal{E}}^{\top}D^{-1/2}$, where $D$, $D_{\mathcal{E}}$, and $W_{\mathcal{E}}$ are diagonal matrices of node degree, hyperedge degree and hyperedge weight, respectively.

Inspired by hypergraph, we further consider a generalization of line graph (the dual graph represents the adjacencies between edges of a graph).
We denote hypernode as a set of arbitrary number of edges of graph $\mathcal{G}=(\mathcal{V},\mathcal{E},\omega)$, $h_{\mathcal{V}} \subset \mathcal{E}$ and its hypernode degree is $\delta_{\mathcal{V}}(h_{\mathcal{V}}) = |h_{\mathcal{V}}|$.
Then the augmented graph with hyperstructure becomes $\mathcal{H}_{\mathcal{G}}=(\mathcal{V},\mathcal{E},\mathcal{H}_{\mathcal{V}},\mathcal{H}_{\mathcal{E}},\omega, \omega_{\mathcal{V}},\omega_{\mathcal{E}})$.
The connectivity between edge and hyperedge can be analogously expressed by the incidence matrix $Q_{\mathcal{V}}$.
Note that node degree matrix $D$ is also updated by corresponding edge degree in the hyper-line-graph, $d(e)=\sum_{h_{\mathcal{V}}}\omega_{\mathcal{V}}(h_{\mathcal{V}})q(e,h_{\mathcal{V}})$. 
So we can utilize message propagation with high-order structures in the GNN framework of \cite{Gilmer2017icml} to capture high-order interdependencies in the data. The detailed process of building hyperstructures for a distribution network can be found in Algorithm~\ref{hyperstructures_search}.

\begin{figure}[h]
\centering
\includegraphics[width=0.47\textwidth]{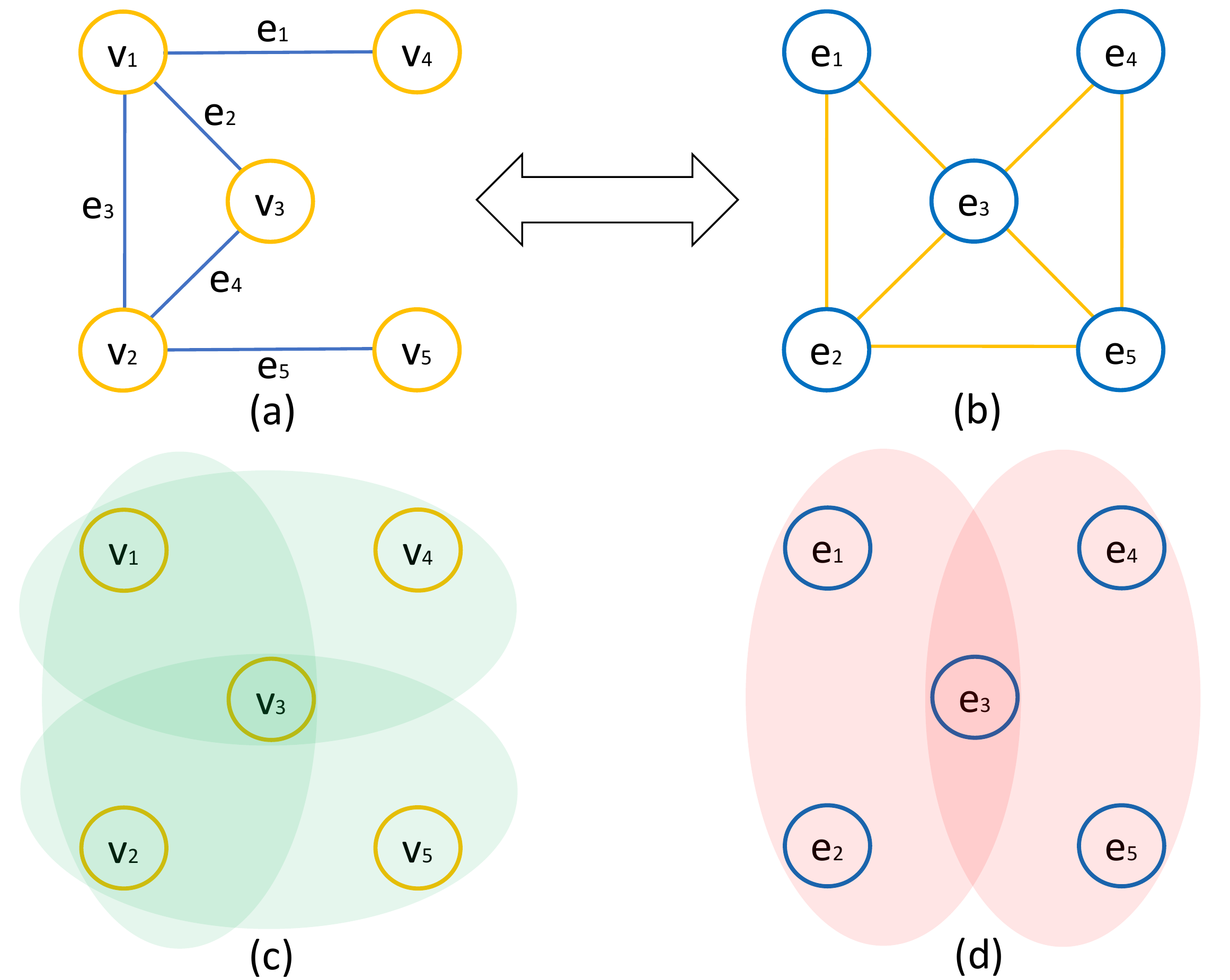}
\caption{An example of graph and hyperstructures: (a) graph , (b) line graph, (c) hypergraph, and (d) hyper-line-graph.
\label{hyperstructure}}
\end{figure} 

\begin{algorithm}
\KwInput{Graph $\mathcal{G} = (\mathcal{V}, \mathcal{E})$}
\KwOutput{Hyperedge $H_{\mathcal{E}}$ and Hypernode $H_{\mathcal{V}}$}
\KwFunction{$k$-nearest neighborhood $f_{knn}$}{
\For{$v$ in substation nodes}{
  $e_h$ = $f_{knn}$($v$, $\mathcal{V}$, $k$)\;
  $H_{\mathcal{E}}$.insert($e_h$)
}
\For{$e$ in edge set $\mathcal{E}$}{
  \If{$e$ is incident to a substation}
{
$v_h$ = $f_{knn}$($e$, $\mathcal{E}$, $k$)\;
$H_{\mathcal{V}}$.insert($v_h$)
}}
\Return{$H_\mathcal{E}, H_\mathcal{V}$}\;
  }
\caption{Hyperstructures Construction\label{hyperstructures_search}}
\end{algorithm}

Fig.~\ref{hyperstructure} presents a toy example of our augmented graph with hyperstructures. 
The top left diagram gives a graph with five nodes and five edges. Its line graph is shown on the right, for example, edge $e_3$ connects all other edges since it shares node $v_1$  with  $e_1$ and $e_2$, and shares node $v_2$ with $e_4$, $e_5$ in the original graph. 
The bottom diagrams illustrate hyperedge about nodes in (c) and hypernode about edges in (d). 
For real data analysis, we may predefine hyperstructures from contexts or build them according to graph structure characteristics and some clustering algorithms.

\subsection{Hyperstructures Graph Convolutional Neural Networks}
Although graph-based semi-supervised learning (G-SSL) yields promising performance on many datasets,
G-SSL uses only the given 
adjacency matrix $A$ and the label matrix $Y$, but not the feature matrix $X_{\mathcal{V}}$. This limitation is important, particularly when classifying data that not only exhibit a sophisticated topological structure but also provide valuable information on node features. To address this limitation, there have been recently proposed many graph-based neural networks methods, e.g., GCN, which use the feature matrix $X_{\mathcal{V}}$ instead of the label matrix $Y$ and encode the graph structure by using neural network framework. Here, we interpret GCN as the learnable graph filter in the spectral domain.

\subsubsection{Graph Convolutional Networks}
GCN conducts graph filtering operation in each convolutional layer with the filter $g$ and the signal matrix $Z_{\textnormal{GC}}^{(\ell-1)}$ (i.e., the matrix of activations fed to the $\ell$-th layer), where $g = I - L$ and the initial input signal matrix $Z_{\textnormal{GC}}^{(0)} = X_{\mathcal{V}}$. Hence, we can obtain the following expression 
$$g_{\theta^{\prime}} \star x_{\mathcal{V}} \approx \theta_{0}^{\prime} x_{\mathcal{V}}+\theta_{1}^{\prime}\left(I-L\right) x_{\mathcal{V}} =\theta_{0}^{\prime} x_{\mathcal{V}} -\theta_{1}^{\prime} D^{-{1}/{2}} A D^{-{1}/{2}} x_{\mathcal{V}}.$$
When constraining a single parameter in this convolutional filter, i.e., $\theta = \theta^\prime_0 = -\theta^\prime_1$,
the filter operation becomes
$$g_{\theta} \star x_{\mathcal{V}}  \approx \theta(I + D^{-{1}/{2}} A D^{-{1}/{2}}) x_{\mathcal{V}}.$$
To avoid the exploding and vanishing gradient, GCN transforms 
$I + D^{-{1}/{2}} A D^{-{1}/{2}}$ to $\tilde{D}^{-{1}/{2}} \tilde{A} \tilde{D}^{-{1}/{2}}$ using the re-normalization trick, where $\Tilde{A} = I + A$ and $\tilde{D}_{i i}=\sum_{j} \tilde{A}_{i j}$.
Finally, the graph convolution operation on node-level representation learning can be formulated as
\begin{align}
\label{gcn_conv}
Z_{\textnormal{GC}}^{(\ell)}=f_{\textnormal{GMP}}(\sigma(\tilde{D}^{-{1}/{2}} \tilde{A} \tilde{D}^{-{1}/{2}} Z_{\textnormal{GC}}^{(\ell-1)} \Theta^{(\ell-1)})),
\end{align}
where $f_{\text{GMP}}$ is the global max pooling, $\sigma(\cdot)$ is the activation function, e.g., ReLU, and $\Theta^{(\ell-1)}$ is the trainable weight matrix in the layer.

\begin{figure}[h]
\centering
\includegraphics[width=0.49\textwidth]{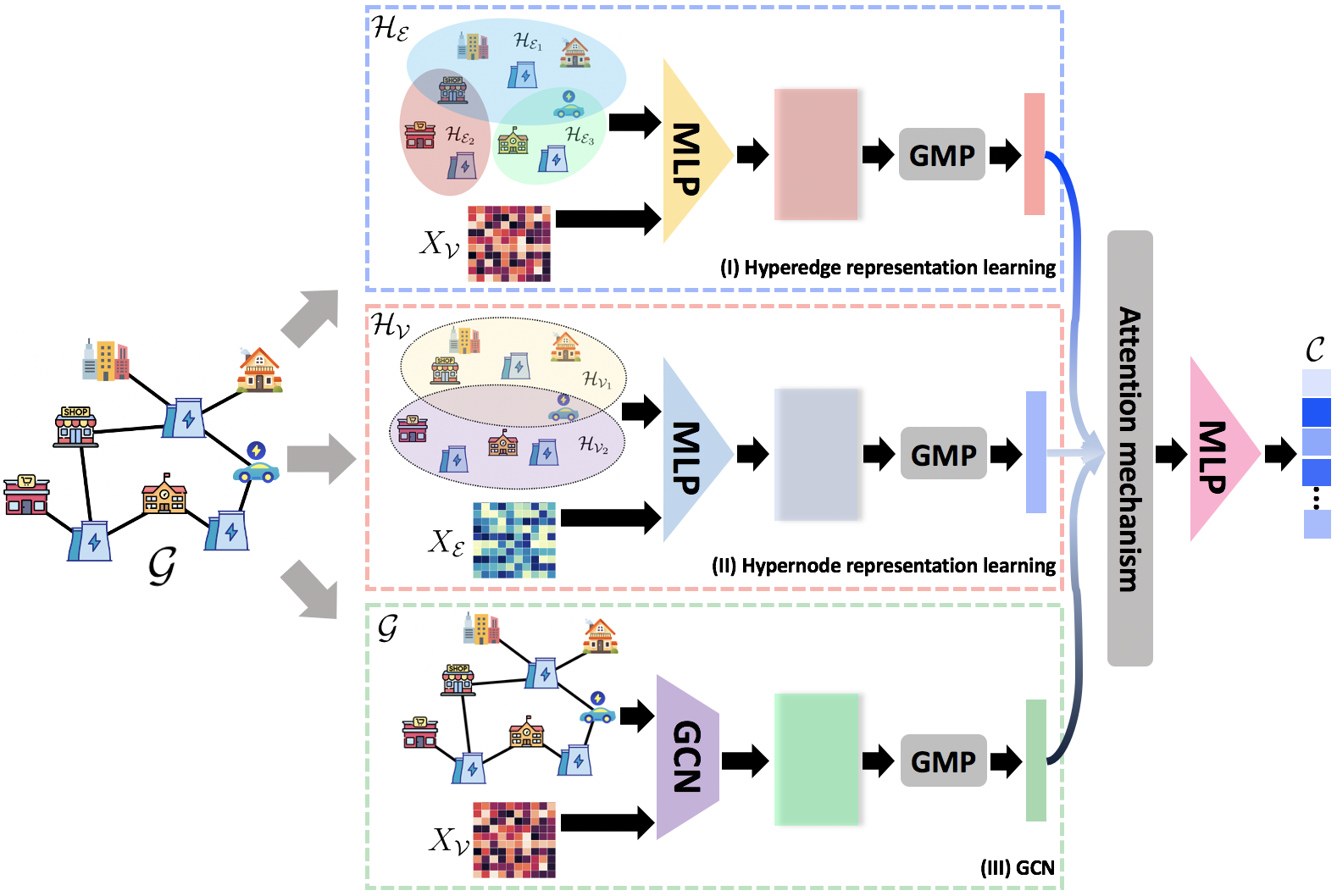}
\caption{The Hyper-GCNNs framework consists of three components: (I) hyperedge representation learning scheme, (II) hypernode representation learning scheme, and (III) graph convolution operation.
\label{hyper_gnets_flowchart}}
\end{figure} 

\begin{table*}[h]
\centering
\caption{Average accuracy (\%) ($\pm$standard error) comparison with different methods on 54-bus system \rom{1} and 54-bus system \rom{2}.\label{overall_result}}
\begin{tabular}{lcccccc}
\toprule
\multirow{2}{*}{\bf Model}& \multicolumn{2}{c}{\bf 3 classes} & \multicolumn{2}{c}{\bf 4 classes}& \multicolumn{2}{c}{\bf 5 classes}
\\
\cmidrule(lr){2-3}\cmidrule(lr){4-5}\cmidrule(lr){6-7}
& 54-bus system \rom{1}  & 54-bus system \rom{2}   & 54-bus system \rom{1}  & 54-bus system \rom{2} & 54-bus system \rom{1}  & 54-bus system \rom{2} \\
\midrule
GCN~\cite{kipf2017semi}&89.10$\pm$1.98 &72.00$\pm$2.39 &84.90$\pm$2.61 &69.14$\pm$2.17 &73.00$\pm$2.32 &65.43$\pm$2.56 \\
GraphSAGE~\cite{hamilton2017inductive}&90.60$\pm$2.46 &82.38$\pm$2.33 &85.00$\pm$2.39 &80.07$\pm$3.89 &72.90$\pm$3.63 &67.45$\pm$4.56\\
GAT~\cite{velivckovic2018graph} &88.70$\pm$2.64 &80.00$\pm$2.80 &84.10$\pm$2.77 &79.96$\pm$3.98 &71.10$\pm$1.71 &69.89$\pm$3.78\\
GIN~\cite{xu2018powerful}&89.10$\pm$2.12 &84.61$\pm$3.28 &85.20$\pm$2.70 &76.60$\pm$2.99 &73.00$\pm$2.63 &66.51$\pm$3.39\\
DiffPool~\cite{ying2018hierarchical}&90.90$\pm$3.39 &85.71$\pm$3.30 &84.33$\pm$2.47 &78.68$\pm$4.00 &75.50$\pm$3.61&73.02$\pm$5.02 \\
SNNs~\cite{ebli2020simplicial}  &  83.80$\pm$2.55 &81.39$\pm$3.76 &79.50$\pm$2.42 &73.95$\pm$3.33 &70.20$\pm$2.95 &63.00$\pm$3.40\\
FC-V~\cite{o2021filtration}&86.80$\pm$2.71 &82.85$\pm$2.15 &81.30$\pm$2.66 &75.56$\pm$2.47 &71.66$\pm$3.55 &64.75$\pm$2.96\\
\midrule 
{\bf Hyper-GCNNs (ours)} &\bf{91.80$\pm$2.34} &\bf{88.28$\pm$3.17} &\bf{86.37$\pm$1.65} &\bf{84.29$\pm$2.26} &\bf{78.00$\pm$2.11} &\bf{73.35$\pm$3.29}\\
\bottomrule
\end{tabular}
\end{table*}

\subsubsection{Hyperstructures Representations Learning}
In our experiments, we use Multilayer Perceptron (MLP) to learn the hyperstructures (i.e., hyperedge and hypernode) of a given distribution system. Given the hyperedge $\mathcal{H}_{\mathcal{E}}$, we employ MLPs and global max pooling (GMP) to obtain the graph-level representation
\begin{align}
    Z_{\mathcal{E}} = f_{\textnormal{GMP}}(\textnormal{MLP}(\mathcal{H}_{\mathcal{E}} \cdot \textnormal{MLP}(\mathcal{H}^{\top}_{\mathcal{E}}X_{\mathcal{V}}))),
\end{align}
where $\textnormal{MLP}$ is a MLP model, $X_{\mathcal{V}} \in \mathbb{R}^{N \times c_{\mathcal{V}}}$ represents the node feature matrix, and $Z_{\mathcal{E}} \in \mathbb{R}^{d^{\textnormal{out}}_{\mathcal{E}}}$ denotes the output embedding for a bus system based on hyperedge information. Similarly, the graph-level representation based on hypernode $\mathcal{H}_{\mathcal{V}}$ can be formulated as
\begin{align}
    Z_{\mathcal{V}} = f_{\textnormal{GMP}}(\textnormal{MLP}(\mathcal{H}_{\mathcal{V}} \cdot \textnormal{MLP}(\mathcal{H}^{\top}_{\mathcal{V}}X_{\mathcal{E}}))),
\end{align}
where $X_{\mathcal{E}}$ is the edge feature matrix and $Z_{\mathcal{V}} \in \mathbb{R}^{d^{\textnormal{out}}_{\mathcal{V}}}$ represents the output embedding for a distribution system based on hypernode information.

To adaptively learn the intrinsic dependencies among different representations from graph structure and hyperstructures, we utilize the attention mechanism to focus on the importance of task relevant parts of the learned representations for decision making, i.e., $(\alpha_{\textnormal{GC}}, \alpha_{\mathcal{E}}, \alpha_{\mathcal{V}}) = Att(Z_{\textnormal{GC}}, Z_{\mathcal{E}}, Z_{\mathcal{V}})$. In practice, we compute the attention coefficient as follows
\begin{align}
\begin{split}
    \alpha_i &= \text{softmax}_i(\Upsilon_{\text{Att}} \tanh{(\Xi Z_i)}) \\
     &= \frac{\exp{(\Upsilon_{\text{Att}} \tanh{(\Xi Z_i)})}}{\sum_{j\in \{\textnormal{GC}, \mathcal{E}, \mathcal{V} \}}\exp{(\Upsilon_{\text{Att}} \tanh{(\Xi Z_j)})}},
\end{split}
\end{align}
where $\Upsilon_{\text{Att}} \in \mathbb{R}^{1 \times d_{\text{out}}}$ is a linear transformation, $\Xi$ is the trainable weight matrix, and the softmax function is utilized to normalize the attention vector. Then we obtain the final embedding $Z$ by combining all embeddings  
\begin{align}
    Z = \alpha_{\textnormal{GC}} \times Z_{\textnormal{GC}} + \alpha_{\mathcal{E}} \times Z_{\mathcal{E}} + \alpha_{\mathcal{V}} \times Z_{\mathcal{V}}.
\end{align}
Lastly, we feed the final embedding $Z$ into an MLP layer and use a differentiable classifier for graph learning tasks. The overall framework of Hyper-GCNNs is shown in Fig.~\ref{hyper_gnets_flowchart}.

\section{Experimental Studies}
\label{exp}
{\bf Datasets} In our experiments, we consider learning CVaR of Annual Load Shedding through multi-class classification and graph regression. For the graph classification tasks, to better study the challenges of CVaR analysis under different expansion plans, we work on three different number of classes based on ranges of annual CVaR of loss of load (kWh), i.e., 3 classes, 4 classes, and 5 classes. Table~\ref{CVaR_label} shows the details of these classes and their respective ranges of annual CVaR of loss of load. {Here we follow a standard statistical practice~\cite{nargesian2017learning} and bin
expansion plans into classes based on quantile ranges of annual CVaR of loss of load (kWh), that is, 3 classes (i.e., low-, moderate-, and high-risk), 4 classes (i.e., low-, moderate-, middle-, and high-risk), and 5 classes (i.e., low-, moderate-, high-, very high-, and extreme high-robustness). Such binning can be viewed, for example, as an analogue of quantitative requirements under Basel III for banking and the regulatory framework of Solvency II for the European insurance industry, where CVaR is referred to as expected shortfall (ES) in financial management and  supervision. Optimal binning practices and their connection to planning standards are topics of standalone interest which we leave for future research.} For example, when using 3 classes for the 54-bus system I, we consider the 1st class as the range between 0 kWh and 10,000 kWh, the 2nd class associated with the range between 10,000 kWh and 20,000 kWh, and the 3rd class related to more than 20,000 kWh of annual CVaR of loss of load.
\begin{table}[h!]
\centering
\setlength\tabcolsep{3pt}
\caption{Ranges of annual CVaR of loss of load.\label{CVaR_label}}
\begin{tabular}{lcccc}
\toprule
\multirow{2}{*}{\textbf{\# Classes}}& \multicolumn{2}{c}{\textbf{54-bus system \rom{1}}} & \multicolumn{2}{c}{\textbf{54-bus system \rom{2}}}
\\
\cmidrule(lr){2-3}\cmidrule(lr){4-5}
& {Label} & {Range (kWh)}& {Label} & {Range(kWh)}\\
\midrule
\multirow{3}{*}{\# 3 classes} & 0 &[0, 1.0e4]&0&[0, 3.0e4]\\
&1&(1.0e4, 2.0e4]&1&(3.0e4, 4.0e4]\\
&2&(2.0e4, $\infty$)&2&(4.0e4, $\infty$]\\
\midrule
\multirow{4}{*}{\# 4 classes} & 0 &[0, 1.0e4]&0&[0, 3.0e4]\\
& 1 &(1.0e4, 2.0e4]&1&(3.0e4, 3.5e4]\\
& 2 &(2.0e4, 3.0e4]&2&(3.5e4, 4.0e4]\\
& 3 &(3.0e4, $\infty$]&3&(4.0e4, $\infty$)\\
\midrule
\multirow{5}{*}{\# 5 classes} & 0 &[0, 1.0e4]&0&[0, 3.0e4]\\
& 1 &(1.0e4, 1.5e4]&1&(3.0e4, 3.5e4]\\
& 2 &(1.5e4, 2.0e4]&2&(3.5e4, 4.0e4]\\
& 3 &(2.0e4, 2.5e4]&3&(4.0e4, 4.5e4]\\
& 4 &(2.5e4, $\infty$)&4&(4.5e4, $\infty$)\\
\bottomrule
\end{tabular}
\end{table}

The proposed methodology has been applied to two distribution systems, namely 54-bus system I and 54-bus system II, which are modified versions of the 54-bus system described by~\cite{Munoz2016}. In the 54-bus system I, we have 72 lines (50 existing and 22 candidate lines respectively), 4 substation nodes and 50 load nodes. In the 54-bus system II, we have 72 lines (52 existing and 20 candidate lines respectively), 2 substations, 50 load nodes and 2 non-load nodes. Node features consists of (i) U-scores (where the dimension of U-scores depends on the number of substations), (ii) node degree, and (iii) node betweenness centrality; thus, the feature matrix of 54-bus system is $X_{\mathcal{V}} \in \mathbb{R}^{N \times (\mathcal{N}_{\textnormal{substation}} + 2)}$. Moreover, we use pairwise cosine similarity of node information as edge features $X_{\mathcal{E}} \in \mathbb{R}^{M \times 1}$ (where $M$ denotes the number of edges).

In our experiments, we have generated several possible expansion plans for the two aforementioned systems (200 for the 54-bus system I and 74 for 54-bus system II). These expansion plans have been generated by selecting different subsets of the available candidate lines. Then, for each expansion plan of each system, we have simulated 2000 scenarios of annual operation, with hourly resolution. {We simulate independent Bernoulli trials for the availability of line segments of the distribution grid for each hour of each scenario, setting the rate of routine failures (single-line failures) as 0.4 times per year and the rate of HILP failures (failures involving more than one line segment) as 0.01 times per year.} As a result, we attain the CVaR of annual loss of load for each expansion plan as described in Section \ref{sec:BaseResilienceEvaluationMethod}. The procedure to obtain the dataset has been implemented using Julia 1.1 via the JuMP package combined with CPLEX 12.9, which is run on Intel Core i5-7200U CPU @2.50GHz. For 54-bus system I dataset, we build hyperedge for each of the 4 substations by 10 nearest neighbors in its feature space; for hypernodes, edges incident to substation nodes are selected as centroids and clustering is based on concatenated feature space of incident nodes. The number of edges vary from 50 to 67 over the 200 grids and the number of hypernodes vary from 8 to 12 based on connectivity of grids. A similar preprocessing is performed on the 54-bus system II dataset with 2 substations. There are 52 to 56 edges and resulted 4 to 5 hypernodes over the 74 distribution grids.

{\bf Baselines} For the graph classification tasks, we compare our Hyper-GCNNs with 7 baselines representative of three different categories (i.e., GNN-based models, simplicial complex-based neural networks, and topological machine learning method), including (i) Graph Convolutional Networks (GCN)~\cite{kipf2017semi}, (ii) GraphSAGE~\cite{hamilton2017inductive}, (iii) Graph Attention Networks (GAT)~\cite{velivckovic2018graph}, (iv) Diffusion-convolutional neural networks (GIN)~\cite{atwood2015diffusion}, (v) DiffPool~\cite{ying2018hierarchical}, (vi) Simplicial Neural Networks (SNNs)~\cite{ebli2020simplicial}, and (vii) Filtration Curves with Random Forest (FC-V)~\cite{o2021filtration}. For the graph regression tasks, we implement the Random Forest, GCN~\cite{kipf2017semi}, and GraphSAGE~\cite{hamilton2017inductive} as baselines. We use the generalization
accuracy and Root Mean Squared Errors (RMSE) to measure the performances on 54-bus system \rom{1} and 54-bus system \rom{2} for graph classification and graph regression respectively. Note that, we split the dataset (i.e., 54-bus system \rom{1} and 54-bus system \rom{2}) to different partitions as training and test sets splits of 80\% and 20\%, respectively. 

{\bf Experimental Settings}
We use the Adam optimizer for 100 epochs to train Hyper-GCNNs. For the graph classification tasks, (i) 54-bus system \rom{1}: Hyper-GCNNs consists of 3 layers whose hidden feature dimension is 64, and each layer consists of 2 MLP blocks; the learning rate is 0.01 the dropout is set as 0.5, and the batch size is set as 16; and (ii) 54-bus system \rom{2}: Hyper-GCNNs consists of 3 layers whose hidden feature dimension is 16, and each layer consists of 2 MLP blocks; the learning rate is 0.05 the dropout is set as 0.0, and the batch size is set as 8. For the graph regression tasks, (i) 54-bus system \rom{1}: Hyper-GCNNs consists of 3 layers whose hidden feature dimension is 64, and each layer consists of 2 MLP blocks; the learning rate is 0.01 the dropout is set as 0.5, and the batch size is set as 8; and (ii) 54-bus system \rom{2}: Hyper-GCNNs consists of 2 layers whose hidden feature dimension is 8, and each layer consists of 2 MLP blocks; the learning rate is 0.0001 the dropout is set as 0.8, and the batch size is set as 8. Moreover, for both datasets, the filter size, kernel size, stride, and the size of global max pooling is set to be 8, 2, 2, and $3\times3$ respectively. In our work, hyperedges and hypernodes are based on 10-nn (i.e., $k=10$ in $k$-nearest neighbor algorithm) about substation nodes and edges incident to substation nodes, respectively. For the 54-Bus system I, we have 4 hyperedges and around 10 hypernodes for each plan; for the 54-Bus system II, we have 2 hyperedges and around 4 hypernodes for each plan. We select the best hyperparameter setting of our Hyper-GCNNs through cross-validation. For baselines, we use the original papers' codes in related code repositories. Our data and codes are publicly available at~\url{https://github.com/hypergcnns/hypergcnns.git}.

{\bf Findings} Graph classification results on two 54 bus systems, averaged over 10 cross-validation runs, are summarized in Table~\ref{overall_result}. It suggests the following key observations: (i) Hyper-GCNNs surpasses all state-of-the-art baselines in terms of
classification performance over all considered scenarios across all datasets; (ii) Hyper-GCNNs brings relative gains with respect to the next best approach from 0.98\% (for 3 classes of the 54-bus system II and DiffPool as competitor) to 3.20\% (for 5 classes of the 54-bus system I and DiffPool as competitor); (iii) Hyper-GCNNs yields substantial reductions in computational costs with respect to the existing stochastic optimization  methods adopted by the power system community and the {\it second} lowest running time among DL models, i.e., the running time (seconds: {\it s}) of the methods tested by us on 54-bus system (i.e., 54-bus system \rom{1}) are: GCN (1.00 {\it s}); GAT (1.31 {\it s}); GraphSAGE (1.55 {\it s}); GIN (1.80 {\it s}); DiffPool (2.03 {\it s}); SNNs (2.50 {\it s}); and FC-V (0.80 {\it s}); and (iv) as expected, performance of all models deteriorates with an increase of a number of classes. The graph regression evaluation results are summarized in Table~\ref{regression_res}. Compared with 3 baseline methods, our Hyper-GCNNs has achieved the best performance on both 54-bus system \rom{1} and 54-bus system \rom{2} datasets. On average, Hyper-GCNNs outperforms the best baseline by 2.5\% on RMSE. These results prove that Hyper-GCNNs to be the most competitive approach for both graph classification and graph regression of the distribution expansion plans 
both in terms of computational costs and accuracy.

\begin{table}[h!]
\centering
\caption{The CVaR prediction performance (RMSE$\pm$standard error) of Hyper-GCNNs and different methods on 54-bus system \rom{1} and 54-bus system \rom{2}.\label{regression_res}}
\begin{tabular}{lcc}
\toprule
\textbf{Model} & \textbf{54-bus system \rom{1}} & \textbf{54-bus system \rom{2}} \\
\hline
Random Forest &0.68$\pm$0.011 &0.94$\pm$0.001 \\
GCN~\cite{kipf2017semi} &0.59$\pm$0.007 &1.10$\pm$0.001 \\
GraphSAGE~\cite{hamilton2017inductive} &0.57$\pm$0.005 & 1.00$\pm$0.001\\
\midrule
{\bf Hyper-GCNNs (ours)} &\bf{0.55$\pm$0.003} &\bf{0.93$\pm$0.001}\\
\bottomrule
\end{tabular}
\end{table}

\begin{table}[h]
\centering
\caption{Computational costs for U-scores generation under each expansion plan and a single training epoch of Hyper-GCNNs.\label{running_time}}
\begin{tabular}{lccc}
\toprule
\multirow{2}{*}{\textbf{Dataset}} & \multirow{2}{*}{\textbf{Avg. \# edges}} & \multicolumn{2}{c}{\textbf{Average Time Taken (sec)}} \\
& & U-scores & Hyper-GCNNs (epoch) \\
\midrule
54-bus system \rom{1}&55.99 &1.81& 1.25\\
54-bus system \rom{2}&54.38 &0.03& 0.89\\
\bottomrule
\end{tabular}
\end{table}

{\bf Ablation Study} we conduct an ablation study to examine the contributions of different components in our proposed Hyper-GCNNs on 54-bus system I (3 classes) and 54-bus system II (3 classes) datasets. That is, we compare our Hyper-GCNNs with four ablated variants, (i) Hyper-GCNNs without U-scores (Hyper-GCNNs w/o U-scores), (ii) Hyper-GCNNs without hyperedge representation learning (Hyper-GCNNs w/o Hyperedge), (iii) Hyper-GCNNs without hypernode representation learning (Hyper-GCNNs w/o Hypernode), and (iv) Hyper-GCNNs without attention mechanism (Hyper-GCNNs w/o Attention mechanism). Table~\ref{ablation_study} shows that, when ablating the above components, the accuracy of Hyper-GCNNs drops significantly on graph classification tasks. These ablation results indicate that integrating the information through U-scores and hyperstructures representations learning with the attention mechanism is critical.
\begin{table}[h!]
\centering
\setlength\tabcolsep{1.pt}
\caption{Ablation study of the architecture of Hyper-GCNNs.\label{ablation_study}}
\begin{tabular}{lcc}
\toprule
\textbf{Architecture} & \textbf{54-bus system \rom{1}} & \textbf{54-bus system \rom{2}} \\
\hline
Hyper-GCNNs &\bf{91.80$\pm$2.34} &\bf{88.28$\pm$3.17} \\
Hyper-GCNNs w/o U-scores &90.40$\pm$2.37 & 86.52$\pm$3.40 \\
Hyper-GCNNs w/o Hyperedge &83.00$\pm$3.15 &81.75$\pm$3.29 \\
Hyper-GCNNs w/o Hypernode &89.00$\pm$2.58 & 84.28$\pm$2.43 \\
Hyper-GCNNs w/o Attention mechanism &90.50$\pm$2.20 & 85.71$\pm$3.28 \\
\bottomrule
\end{tabular}
\end{table}

{\bf Computational Complexity} The computational complexity for the computation of a simple path between load and substation is $\mathcal{O}(N + M)$, where $N$ is the number of nodes and $M$ is the number of edges. Since we consider the U-scores for all nodes in the distribution network, the computational complexity for computation of U-scores is $\mathcal{O}(N^2 + NM)$. 
For $k$-nn hyperstructure generation, the complexity is $\mathcal{O}(N\log(N)+ M\log(M))$ since number of clusters and feature dimensions are usually very small with respect to the numbers of nodes and edges.

\section{Utility and Limitations
}
The purpose of this paper is to investigate the ability of GNN-based model to reproduce risk-based metrics used in electricity distribution grid resilience planning, that is, the area being still an uncharted territory for AI. Our findings indicate that the proposed GNN-based model can outperform the more conventional methods in terms of both accuracy and computational efficiency in the classification of CVaR for loss of load in 
electricity distribution grid expansion plans. Indeed, the computational time to classify different grid plans (in the order of seconds) is encouraging, especially when compared with existing simulation based CVaR evaluation methodologies that may take several hours to compute. This clearly indicates that classification methods, based on GNN architectures, demonstrate a high potential to be successfully incorporated into the traditional expansion and planning algorithms to improve the computational times of the existing large-scale stochastic optimization models. Additionally, the U-scores metric has proved to be an excellent topology-based metric to express the risk of the system. This deserves to be further investigated in future works, as the use of this metric can go beyond planning applications. This is particularly relevant in distribution systems, that are operated radially and where power and voltage stability elements (substations, electric storage resources) are well identified in the network. For example, in these systems, U-scores could be used to improve operations, through the development of methods for fast topological reconfiguration based on simple uniqueness score classifications.

{\bf Not Only Power Distribution Systems} 
Our proposed methodology can be broadly applied way
beyond distribution systems
to various types of networks 
characterized by the prominent multi-node (i.e., higher-order) interactions and heterogeneity of node roles in the underlying system organization such as molecular, transportation, and blockchain transaction graphs. For instance, in molecular networks, the relationships between different cell type markers (e.g., microglia, astrocytes, oligodendrocytes and neurons) can be quantitatively evaluated through U-scores, while hyperstructures representation learning can help to capture hidden dependencies among them - which will result in enhancing treatment outcomes. Similarly, to improve traffic forecasting performance, 
hyperstructures and U-scores can be used to capture the complex spatio-temporal dependencies in human mobility among transportation hubs and rural areas.
Our Hyper-GCNNs approach makes an important step to address such intrinsic network properties in a systematic graph-theoretic manner. 

\section{Discussion}
Distributions networks are responsible for delivering power to individual consumers and are composed by a large number of nodes with small distances between them, resulting in relatively standardized electrical installations under a substation, i.e., at the neighborhood/town level. Large electric utilities have hundreds of such distribution substations, which means that their grid is actually formed by multiple networks with similar properties. This unique aspect of distribution grids opens up possibilities to apply automatic classification algorithms based on GNNs 
to improve power system computations, namely, stochastic planning methods (in the context of resilience) that entail significant computational costs. 

The first step to unlock these applications was achieved with this paper, i.e., showing that it is possible to capture resilience via automatic classification methodologies (which has been achieved with this paper). After this first step, there are two possible ways of including classification algorithms in power grid resilience planning. The first is as a replacement of the traditional resilience and reliability evaluation methodologies, based on SMCS described in Methods Section. In this case, the classification methods based on GNNs are an alternative form to obtain a fast and standardized resilience analysis of the network, i.e., exclusively based on the topology and the grid assets. A tool with these characteristics could be used, for example, by regulators to approve reliability/resilience investments and to compare grid assets or benchmark resilience performance across different territories and utilities. Second, classification methods based on graph convolution operation and hyperstructures representation learning framework can be used as a pre-solver of the stochastic expansion and planning optimization methods. For example, U-scores can be applied in this pre-solver to eliminate potential candidate plans or to identify specific nodes for investments in batteries and substations. This will result in significant decreases of the search space in optimization, reducing computational times and allowing optimal expansion and planning methods to be applied in large-scale distribution systems. 
\section*{Acknowledgment}
This work was authored in part by the Lawrence Berkeley National Laboratory (LBL), U.S. Department of Energy (DOE) National Laboratory operated by the University of California and also was supported by the NSF grant \# ECCS 2039701 and ONR grant \# N00014-21-1-2530. The views expressed in the article do not necessarily represent the views of the DOE, NSF, and ONR. 

\bibliography{reference_HyperGNETs}
\bibliographystyle{IEEEtran}
\end{document}